\documentclass[journal]{IEEEtran}
\usepackage{times}
\usepackage{float}
\usepackage{epsfig}
\usepackage{graphicx}
\usepackage{amsmath}
\usepackage{amssymb}
\usepackage{tabularx}
\usepackage{gensymb,nicefrac}
\usepackage{array,multirow}
\usepackage{url}
\usepackage{tabularx}
\usepackage{booktabs}
\usepackage[normalem]{ulem}

\usepackage{academicons}
\usepackage[switch]{lineno}
\usepackage[table]{xcolor}

\usepackage[pagebackref=true,breaklinks=true,letterpaper=true,colorlinks,bookmarks=false]{hyperref}

\newcommand{\MODEL}{{MOCCA}}
\newcommand{\MODELEXT}{{Multi-layer One-Class ClassificAtion}}

\newcommand{\cifar}{{CIFAR10}}
\newcommand{\mvtec}{{MVTec AD~\cite{bergmann2019mvtec}}}
\newcommand{\shanghai}{{ShanghaiTech~\cite{luo2017revisit}}}
\newcommand{\nasymb}{{-}}

\ifCLASSINFOpdf
\else
\fi

\usepackage{dblfloatfix}
\usepackage{cite}

\usepackage{tikz}
\usepackage{textcomp}

\newcommand\copyrighttext{%
  \footnotesize \textcopyright 2021 IEEE. Personal use of this material is permitted. Permission from IEEE must be obtained for all other uses, in any current or future media, including reprinting/republishing this material for advertising or promotional purposes, creating new collective works, for resale or redistribution to servers or lists, or reuse of any copyrighted component of this work in other works.}
\newcommand\copyrightnotice{%
\begin{tikzpicture}[remember picture,overlay]
\node[anchor=south,yshift=10pt] at (current page.south) {\fbox{\parbox{\dimexpr\textwidth-\fboxsep-\fboxrule\relax}{\copyrighttext}}};
\end{tikzpicture}%
}

\begin{document}



%
\title{\MODEL{}: \MODELEXT{} for Anomaly Detection}
%
%
%

\author{
Fabio Valerio
Massoli\href{https://orcid.org/0000-0001-6447-1301}{\includegraphics[scale=0.5]{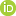}},
\thanks{Fabio Valerio Massoli, Fabrizio Falchi, and Giuseppe Amato are with the Institute of Information Science and Technologies (ISTI) - CNR, Pisa, 56124 Italy (e-mail: \{fabio.massoli; fabrizio.falchi; giuseppe.amato\}@isti.cnr.it).}
Fabrizio Falchi\href{https://orcid.org/0000-0001-6258-5313}{\includegraphics[scale=0.5]{images/orcid_icon.png}},
Alperen Kantarci\href{https://orcid.org/0000-0002-4080-5538}{\includegraphics[scale=0.5]{images/orcid_icon.png}},
\c{S}eymanur Akti\href{https://orcid.org/0000-0003-1778-2136}{\includegraphics[scale=0.5]{images/orcid_icon.png}},
Hazim Kemal Ekenel\href{https://orcid.org/0000-0003-3697-8548}{\includegraphics[scale=0.5]{images/orcid_icon.png}},
\thanks{Hazim Kemal Ekenel, Alperen Kantarci, and \c{S}eymanur Akti are with the Department of Computer Engineering, Istanbul Technical University, Istanbul, Turkey (e-mail: \{ekenel; kantarcia; akti15\}@itu.edu.tr)} 
Giuseppe Amato\href{https://orcid.org/0000-0003-0171-4315}{\includegraphics[scale=0.5]{images/orcid_icon.png}}
}

\maketitle

\copyrightnotice

\begin{abstract}

Anomalies are ubiquitous in all scientific fields and can express an unexpected event due to incomplete knowledge about the data distribution or an unknown process that suddenly comes into play and distorts the observations. 
Usually, due to such events' rarity, to train deep learning models on the Anomaly Detection (AD) task, scientists only rely on ``normal" data, i.e., non-anomalous samples. Thus, letting the neural network infer the distribution beneath the input data. 
In such a context, we propose a novel framework, named \MODELEXT{} (\MODEL{}), to train and test deep learning models on the AD task. 
Specifically, we applied our approach to autoencoders. A key novelty in our work stems from the explicit optimization of the intermediate representations for the task at hand. Indeed, differently from commonly used approaches that consider a neural network as a single computational block, i.e., using the output of the last layer only, \MODEL{} explicitly leverages the multi-layer structure of deep architectures. Each layer's feature space is optimized for AD during training, while in the test phase, the deep representations extracted from the trained layers are combined to detect anomalies.
With \MODEL{}, we split the training process into two steps. First, the autoencoder is trained on the reconstruction task only. Then, we only retain the encoder tasked with minimizing the $L_2$ distance between the output representation and a reference point, the anomaly-free training data centroid, at each considered layer. 
Subsequently, we combine the deep features extracted at the various trained layers of the encoder model to detect anomalies at inference time.
To assess the performance of the models trained with \MODEL{}, we conduct extensive experiments on publicly available datasets, namely CIFAR10, MVTec AD, and ShanghaiTech. We show that our proposed method reaches comparable or superior performance to state-of-the-art approaches available in the literature.
Finally, we provide a model analysis to give insights regarding the benefits of our training procedure.

\end{abstract}

\begin{IEEEkeywords}
Anomaly Detection, One-Class Classification, Deep Learning
\end{IEEEkeywords}

\renewcommand{\b}{\bfseries}
\renewcommand{\u}[1]{\underline{#1}}
\newcommand{\m}[1]{{\scriptsize #1}}
\newcolumntype{C}{>{\centering\arraybackslash}X}
\newcolumntype{M}[1]{>{\centering\arraybackslash}m{#1}}

\section{Introduction} \label{introduction}

\IEEEPARstart{A}{nomalies} represent a controversial phenomenon in the scientific world. Although they can lead to fascinating discoveries, sometimes they are a symptom of something unexpected that just happened. Even though they can manifest in different ways, all kinds of anomalies origin from a common basic principle: an unexpected prediction from a given theory from what is believed to be a proper answer. 

\begin{figure}[t]
\includegraphics[width=\linewidth]{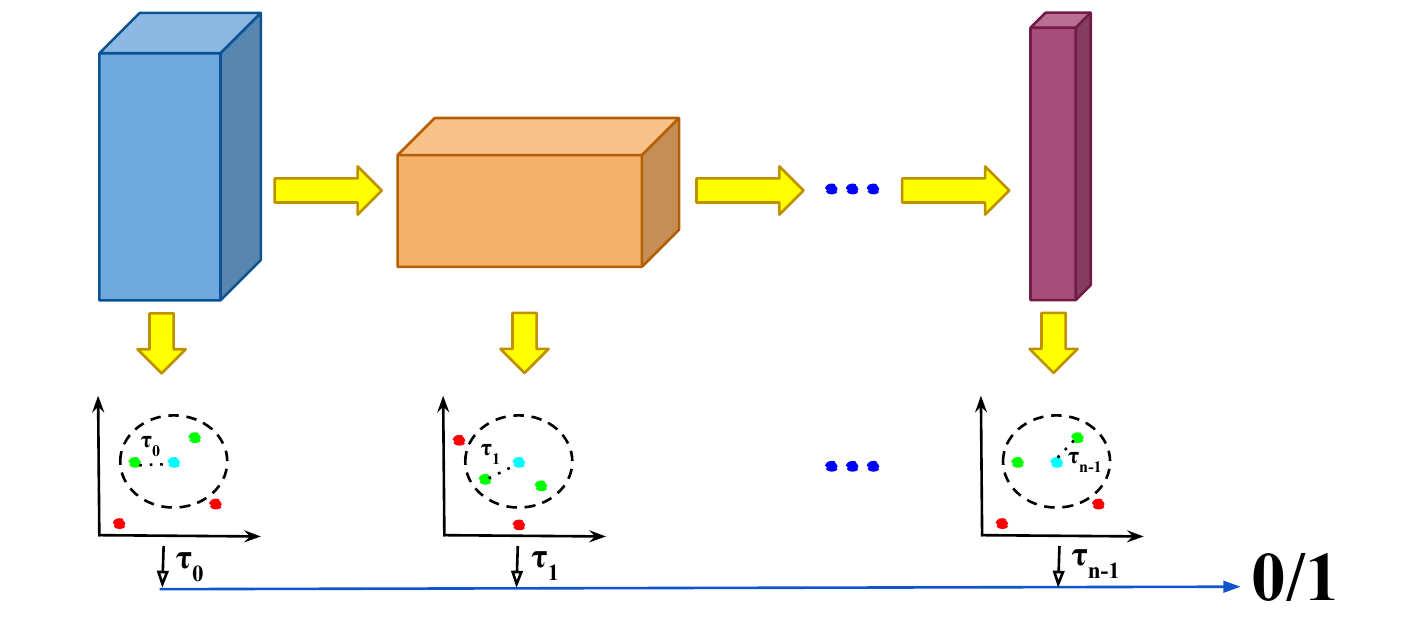}
\caption{Schematic representation of the \MODEL{} approach. Each feature space is represented as an $x-y$ plane. The cyan dots represent the centroids of the anomaly-free images while the green (red) dots represent the normal (anomalous) samples, respectively. $\tau_i$ is the distance between the deep representation of a given input image from the centroid at layer $i$.} \label{fig:arch_simpl}
\end{figure}

Concerning the Deep Learning (DL) field, an anomaly might be thought of as an out-of-distribution sample presented as input to a Deep Neural Network (DNN). More specifically, from a statistical point of view~\cite{hawkins1980identification, edgeworth1887xli}, we can discern among outliers and novelties that are described by the same probability distribution of the normal data and anomalies that are instead characterized by completely different statistics. Being able to detect such events is an attractive feature, especially concerning applications such as surveillance systems~\cite{vinayakumar2019deep, zavrak2020anomaly, alhakami2019network, hasan2016learning}, medical diagnosis~\cite{stafford1994application,fernando2020neural,ouardini2019towards,schlegl2017unsupervised}, fraud detection~\cite{roy2018deep,pumsirirat2018credit,lebichot2019deep,fiore2019using}, and defect detection~\cite{tout2017automatic,kumar2008computer}. Indeed the task of Anomaly Detection (AD)~\cite{aggarwal2015outlier,chalapathy2019deep} is among the most active research fields in the machine learning community.

Since the cost to collect large amounts of anomalous samples is prohibitive, the AD is usually considered as an unsupervised problem with the training databases containing non-anomalous class instances only. Thus, to detect anomalies, deep models are typically trained on in-manifold samples only to learn an effective boundary that captures the concept of normality from the distribution of one kind of data only. In recent years, One-Class (OC) approaches to AD have drawn the scientific community's interest. Especially, autoencoders~\cite{zhou2017anomaly,gong2019memorizing,wang2020advae} and GANs~\cite{schlegl2019f,akcay2018ganomaly,li2019mad} based approaches reached the highest performance available in the literature.

In the Machine Learning (ML) field, commonly adopted approaches leverage the models' final output only, thus interpreting a neural network as a single computational block that performs an input-to-output mapping. 
Concerning such a point of view, throughout this manuscript, we refer to such an approach as ``holistic" interpretation. Specifically, what we mean by ``holistic" is that both the training and test phases rely on the output of the last layer only, i.e., there is no information extracted from the intermediate levels of the architecture.

In such a context, our contribution stems from a different interpretation of the mapping represented by a DNN. We show that by leveraging the deep representations extracted at various depths in both the training and inference phases of a learning model, a neural network reaches higher performance on the AD task than when only the last layer's output is considered. 
We propose a novel framework, named \MODELEXT{} (\MODEL{}), to train and test deep learning models on the AD task. The innovation in our work is the explicit optimization of the intermediate representations and their use in the test phase for the task at hand. 
\MODEL{} leverages the multi-layer structure of deep architectures, differently from commonly used approaches that consider a neural network as a single computational block, i.e., using the output of the last layer only. 
During training, each layer's feature space is optimized for AD, whereas in the test phase, the deep representations extracted from the trained layers are combined to detect anomalies.
To prove the effectiveness of our strategy, we apply it to autoencoders. Specifically, with \MODEL{}, we split the training process into two steps. First, the autoencoder is trained on the reconstruction task only. Then, we only retain the encoder tasked with minimizing the $L_2$ distance between the output representation and a reference point, the anomaly-free training data centroid, at each considered layer. 
Subsequently, we combine the deep features extracted at the various trained layers of the encoder model to detect anomalies at inference time. We show a schematic view of our approach in \autoref{fig:arch_simpl}. Our contributions can be summarized as follows:
\begin{itemize}
\item we formulate a ``multi-layer'' based approach to AD, named \MODEL{}, that explicitly optimizes the representations extracted at different layers of a deep learning model during training, and then combines them in the test phase to detect anomalies;

\item we perform extensive experiments on publicly available single-image AD datasets, namely, \cifar{} and \mvtec{}, and empirically show that models trained with the \MODEL{} approach reach higher performance compared to the state-of-the-art;

\item we perform experiments on the \shanghai{} dataset, and show that, even though our method is not tailored for video-based AD, it delivers models with performance comparable to state-of-the-art approaches specially designed for such a task. Thus, showing the high generalization capability of our technique;

\item we perform a model analysis to give insights into how our approach works and empirically analyze the benefits of exploiting the representations generated at different layers of a learning model.
\end{itemize}
 
The remainder of the paper is organized as follows.
In \autoref{rel_works}, we briefly review the related works, while in \autoref{approach}, we describe our approach to the anomaly detection task. In \autoref{datasets} and \autoref{exp_results}, we present the datasets we used and report the obtained results on them, respectively.
In \autoref{model_analysis} we perform an analysis of the models, and, finally, in \autoref{conclusions}, we conclude the paper.

\section{Related works} \label{rel_works}

\begin{figure*}[t]
\includegraphics[width=\linewidth]{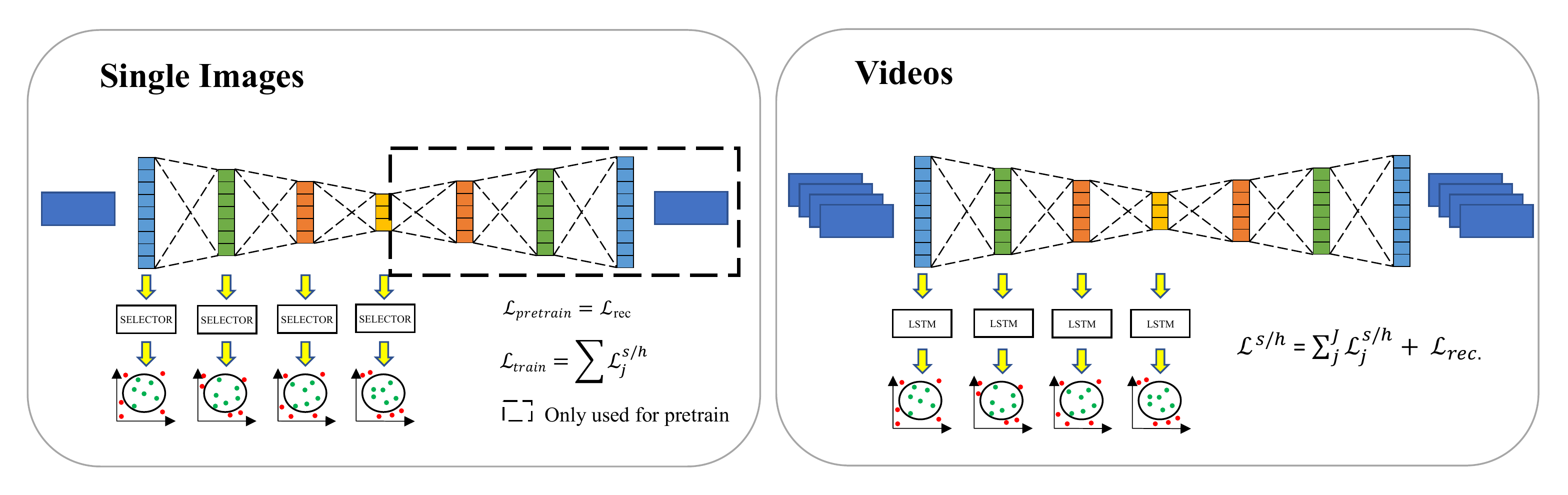}
\caption{Schematic representation of the \MODEL{} training. Left: two-stage training for single image input. Right: end-to-end training for video-based AD. To exploit the time correlation among frames, LSTMs are used instead of ``selector" modules. The superscript $s$ ($h$) refers to the \emph{soft} (\emph{hard}) boundary settings.} \label{fig:arch}
\end{figure*}

The latest approaches to the AD task are mainly based on reconstruction and discrimination techniques. 
Autoencoders~\cite{borghesi2019semisupervised,pawlowski2018unsupervised,li2019specae,zhou2017anomaly} and GANs~\cite{zhou2020sparse,ngo2019fence,sabokrou2018adversarially} belong to the former class while the latter approach gathers techniques such as the one-class classification~\cite{wang2019gods,hong2020latent,razzak2020one}.

Concerning GAN-based approaches, in~\cite{sabokrou2018adversarially}, the authors exploit a reconstruction technique that leverages an autoencoder and a CNN that are adversarially trained. In AnoGAN~\cite{schlegl2017unsupervised} the generator learns to reconstruct the input sample through latent space optimization, and the discriminator generates deep representations for both the original and the reconstructed samples, while in~\cite{zenati2018efficient}, the authors propose to learn an encoder network that maps the input samples directly to the generator's latent space. 
A slightly different approach is 
proposed in~\cite{akcay2018ganomaly}, where an explicit latent space minimization is obtained by learning an encoder model. 
The OC-GAN approach is introduced in~\cite{perera2019ocgan}, where authors use a denoising autoencoder network and a classifier in order to learn the latent representations of the normal samples in an adversarial manner. 

In~\cite{an2015vaead} variational autoencoders are used to detect anomalies by exploiting the reconstruction probability as the objective. 
In~\cite{abati2019latentspace} the authors combine a reconstruction approach based on autoencoders with an autoregressive model that learns a factorization of the latent space distribution.
In~\cite{bergmann2018improving}, the authors use the structural similarity index metric (SSIM) to train 
autoencoders 
while~\cite{huang2019inverse} propose the Inverse-Transform AutoEncoder (ITAE) based on the use of autoencoders that reconstruct images after the application of a set of specific transformations. 

The One-Class (OC) approach has a long history starting from the study of shallow models. Indeed, first attempts in such a direction date back to the 2000s with the proposal of the One-Class SVM~\cite{scholkopf2000ocsvm,tax2004svdd}. 
In~\cite{erfani2016high}, a hybrid approach is proposed based on deep autoencoders and OC-SVM, while in~\cite{chalapathy2018ocnn} the authors trained their models with an OC-SVM equivalent loss function.
One of the first proposals concerning an end-to-end training approach to OC-AD is proposed in~\cite{ruff2018deep}, where the code generated by an encoder is mapped to a point within a hypersphere so that the normal samples remained inside of it while anomalous ones lay outside.
Lastly, in~\cite{oza2019occnn}, the authors use an encoder for getting the latent representations of the normal samples, and a pseudo-negative class is created using zero-centered Gaussian noise in the same latent space. 

Most recently, Venkataramanan et al.~\cite{venkataramanan2020attention} exploit a variational autoencoder combined with a specialized attention mechanism with the final goal of performing anomaly localization. In~\cite{zaheer2020old}, the authors tackle the problem of the stability training of GANs when there are not lots of data available. 
A semi-supervised approach is proposed in~\cite{ruff2019deep}, and 
in~\cite{bergmann2020uninformed}, the authors exploit a student–teacher framework to perform anomaly detection and pixel-precise anomaly segmentation at the same time. 
In \cite{sultani2018real}, they leverage a multiple instance learning approach while in \cite{gong2019memorizing} a technique named MemAE is introduced where authors have added a memory module to deep autoencoders.

Compared to all the works mentioned above, \MODEL{} differs from them on two key aspects. On the one side, it exploits the deep representation extracted at various layers of the learning model, both at training and inference time, which contrasts to classical methodology in which only the final output is considered to fulfill the task. On the other hand, it does not make any assumption on the deep features' statistical distributions. Combining these two properties allows the model to adjust each single feature space at its best to accomplish the AD task. 

\section{Proposed approach} \label{approach}

As a general conception, DNNs are a sequence of transformations that approximate a function $f_{\theta}:\ \mathcal{X} \rightarrow \mathcal{Y}$ where $\mathcal{X} \subseteq \mathbb{R}^d$ and $\mathcal{Y} \subseteq \mathbb{R}^m$ are the input and output space, respectively, and $\theta$ are the parameters to be learned at training time. We refer to such an approach as ``holistic" (see \autoref{introduction} for more details) in the sense that the entire net is considered as a single computational block that given an input, returns an output. As opposed to such a point of view, with \MODEL{} we adopt a ``multi-layer" interpretation of the learning models where we consider a DNN as a sequence of single transformations each mapping its input to a more representative space:

\begin{flalign} \label{eq:layerwise}
\resizebox{1.\hsize}{!}{$f_{\theta}(\mathbf{x}) = \phi_m(\mathbf{\theta}_m; \mathbf{o}_{m-1}) \circ \phi_{m-1}(\mathbf{\theta}_{m-1}; \mathbf{o}_{m-2}) \circ .... \circ \phi_1(\mathbf{\theta}_1; \mathbf{x})$}
\end{flalign}

where each $\phi_i$ term represents the operation performed by a specific layer, and the matrices $\mathbf{\theta}_i$ represent their weights and biases. The output of each operation is reported as $\mathbf{o}_i$, while $\mathbf{x}$ is the network input.

Our intuition is that the outputs $\mathbf{o}_i$ of the various layers, i.e., the representations generated at different depths of a DNN, can be exploited to enhance the performance of a learning model on the AD task compared to when the entire decision process leverages the last layer output only. 
Indeed, it has been already shown in literature~\cite{massoli2020detection,papernot2018deep,carrara2019adversarial} that deep features extracted at various layers of a model can help a DNN to fulfill its task. However, it is not enough to combine the representations at test time only. Instead, all the layers must be trained to a common aim.

As mentioned in \autoref{introduction}, our base network is an autoencoder where both the encoder and the decoder are Deep Convolutional Neural Network (DCNN). With \MODEL{} we formulate the training process as a two-stage procedure in which we first train the full autoencoder on the reconstruction task only, and then we specialize only the encoder to detect anomalies by exploiting an OC-like objective~\cite{ruff2018deep} applied to different layers of the network. However, we empirically observe that a single-step end-to-end training, in which we optimize the reconstruction and the OC objectives simultaneously, is more effective than the two-step one for video-based AD.
A schematic representation of the \MODEL{} training procedures is presented in \autoref{fig:arch}. 
As one can see from the figure, we process the model inner layers' output using ``selector'' and ``LSTM'' modules concerning single-image and video-based data type, respectively. Concerning the ``selector'' blocks, they are made of an average pooling operation or a two-layer neural network concerning the \cifar{} and \mvtec{}, respectively. Specifically, concerning the \cifar{} dataset, we use only the pooling operation to fully assess the real advantages brought by \MODEL{}.

As mentioned above, we exploit the OC objective and we evaluate it by using the deep features extracted at different depths of the encoder model. 
Specifically, we considered two variants for such an objective function termed \emph{soft-} and \emph{hard-}boundary. 
The first one is expressed as follows:

\begin{align} \label{eq:obj_layer_s}
\mathcal{L}_{j}^{s} = R_j^2 + \frac{1}{|B|\cdot\nu}\sum^{|B|}_{i} \mathrm{max}\{0, \parallel \phi_j(\mathbf{x}_i; \theta) - \mathbf{c}_j \parallel^2 - R_j^2\}
\end{align}

The goal of such a loss is to minimize the volume of the hypersphere at each layer $j$, centered at $\bf c_j$ and with radius $R_j$, that is interpreted as the boundary region for normal data~\cite{tax2004svdd}. Then, the goal of~\autoref{eq:obj_layer_s} is to minimize the radius, $R_j$, of such spheres (one for each trained layer). In other words, we expect the ``normal'' data to lie within a sphere, at each layer, while the anomalous samples are expected to remain outside of it. The second addend in the equation penalizes ``normal'' data points that lie outside the sphere after being passed through the network. 
The radius $R_j$ is a scalar quantity evaluated as the $1-\nu$ quantile of the features' distance distribution, in a mini-batch, from the centroid $\bf c_j$. We re-evaluate the radius at each layer at regular intervals while training. A decreasing value of the radius at each layer is an indicator of converging training. The other terms in the equation have the following interpretation: $|B|$ is the mini-batch size, $\nu$ is a hyperparameter that allows controlling the fraction of allowed outliers, $\phi_j$ represents the function that the layer $j$ carries out, and $\mathbf{x}_i$ is the model input. 

Concerning the \emph{hard-}boundary loss, it is expressed as follows:

\begin{align} \label{eq:obj_layer_h}
\mathcal{L}_{j}^{h} &= \frac{1}{|B|\cdot\nu}\sum^{|B|}_{i} \parallel \phi_j(\mathbf{x}_i; \theta) - \mathbf{c}_j \parallel^2
\end{align}

Differently from~\autoref{eq:obj_layer_s}, ~\autoref{eq:obj_layer_h} simply tries to reduce as much as possible the distance of each sample from the layer's centroid by employing a quadratic loss.

After the first training step in which we tasked the full autoencoder with the reconstruction objective, we retain only the encoder and perform an initial forward step on the whole training dataset (that contains non-anomalous samples only) to extract deep features at different depths. Subsequently, we evaluate the centroids, at each layer, as the average of those features. We performed experiments in which we tested the hypothesis of using medoids instead of centroids, but we did not observe any improvement. Once we evaluate the centroids, they are kept fix while training the encoder. We also experimented with several strategies to re-evaluate them after a specific number of training iterations, but we did not observe tangible improvements. 
Regarding the video-based AD, we initialize the centroids at the beginning of the training, i.e., with the model not trained.

Considering a set of layers $\mathcal{J} = \{ j\ |\ j=\ 0,\ 1,\ ... J\}$, we formalize the \MODEL{} objective, during the second-step of the training, as:

\begin{flalign} \label{eq:obj_full}
\mathcal{L}^{s/h} &= \frac{1}{|\mathcal{J}|}\sum_{j}^{|J|}\mathcal{L}_{j}^{s/h} + \frac{\lambda}{2}\sum_{p}^{|P|}\parallel \theta_p \parallel^2 
\end{flalign}

where $|J|$ is the number of layers we consider, and the sum runs over the layer indexes $j$. The last term of the objective is the $L_2$ regularization for the model parameters. 

\section{Datasets and Training} \label{datasets}

This section reports the used datasets and provides details about the training procedure that we adopt.

\subsection{CIFAR10}
The \cifar{} dataset contains 50K training images and 10K test ones shared among ten different classes. We preprocess the images by applying a global contrast normalization procedure using the $L_1$ norm, and then we normalize them to be in the range $[0, +1]$. Given each class, which we refer to as the ``normal class", we have 5000 images to train the model, and we evaluate each model's performance on the whole test set. With such a training approach, the model only sees instances from the ``normal class" and never sees any anomaly while learning.

\subsection{MVTec}\label{mvtecpreprocess}
\begin{figure}[!ht]
\includegraphics[width=\linewidth]{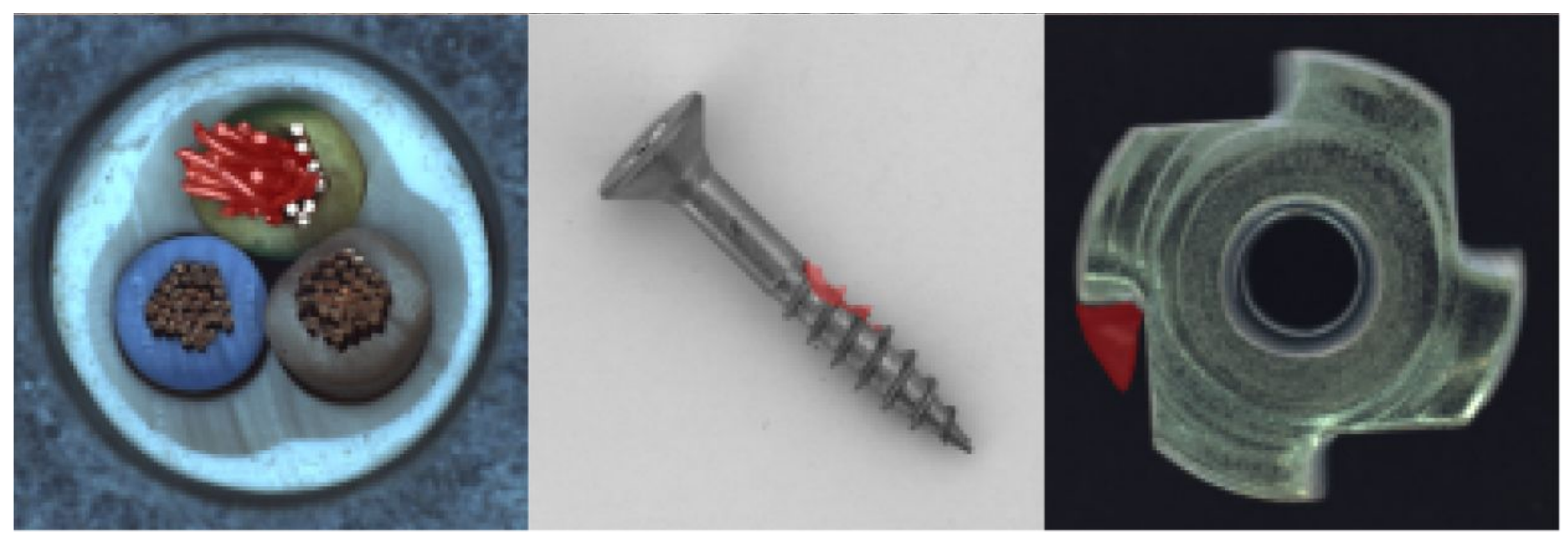}
\caption{Samples from different classes of the \mvtec{} dataset. Top: texture classes. Bottom: object classes. We highlight in red the anomalies.} \label{fig:mvtec_example}
\end{figure}

The \mvtec{} dataset comprises $\sim$3.6K and $\sim$1.7K high-resolution images to train and test DNNs, respectively, shared among 15 classes which are divided into two categories: textures (5 classes) and objects (10 classes). The dataset is split into two sets: one for training purposes containing ``normal" images only and one specifically designed to test the models' performance. Specifically, the latter one contains anomalous images, with defects of different types and non-anomalous ones. We apply two different preprocessing operations to objects- and texture-type classes. Concerning the formers, we first resize the image to 
128x128 pixels and then apply a random rotation in the range $[-\pi/4, +\pi/4]$ when the anomaly of the object is not related to its orientation. Instead, we first resize images to 512x512 pixels in the latter type of classes, and then we crop 64x64 non-overlapping patches used as input to the network. Moreover, we augment the data by exploiting a random rotation in the range $[0, +\pi/4]$. Finally, we normalize all the objects- and texture-type images to be in the range $[-1, +1]$. In \autoref{fig:mvtec_example} we show an example of textures- and objects-type images from the  dataset.

\subsection{ShanghaiTech}
\begin{figure}[!ht]
\includegraphics[width=\linewidth]{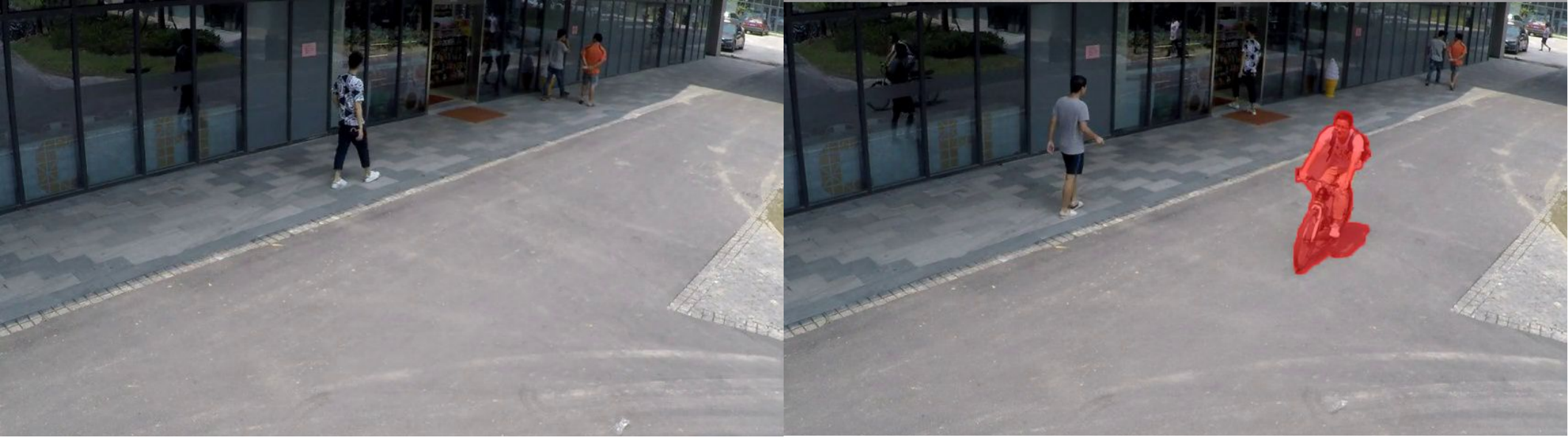}
\caption{Samples of ``normal" (left) and anomalous (right) frames from the \shanghai{} dataset. We highlight in red the anomalies.} \label{fig:shanghaitect_example}
\end{figure}

The \shanghai{} dataset is one of the largest video anomaly datasets. It comprises over 270,000 training frames from 13 scenes with complex light conditions and camera angles, accounting for 130 abnormal events. 
We follow the same preprocessing strategy as in~\cite{abati2019latentspace}, i.e., we use a MOG-based approach to estimate the background and remove it from the frames. By employing such a procedure, we eliminate the necessity of background estimation and let the model focus on foreground objects only. Given a video, we construct clips made by 16 frames to be used as input to the learning models. To exploit the temporal correlation among frames, we employ LSTM cells (we refer the reader to \autoref{approach} for more details about our models' architecture). Finally, we resize each frame to 256x512 pixels to feed models.  In \autoref{fig:shanghaitect_example} we report an example of ``normal" and anomalous frames from two different videos.

\subsection{Training details}\label{train_details}
Concerning the \cifar{} dataset, we use a LeNet-like architecture as in~\cite{ruff2018deep}, made of three convolutional layers and one fully connected layer after them.
We use the Adam~\cite{kingma2014adam} optimizer for both pre-train the full architecture and train the encoder with learning rates of $1.e^{-3}$ and $1.e^{-4}$, respectively. We set the encoder code's size equals to 128 and the value of the parameter $\nu$ in the range [0, 0.1]. Finally, we use a batch size of 256. As we mentioned in \autoref{approach}, concerning the \cifar{} dataset, we use an average pooling operation as the ``selector" module. Thus, we emphasize that the higher performance reached by using \MODEL{} is not due to larger, deeper, or more models. Instead, the benefits of using \MODEL{} stand from its ability to exploit the representations generated at different depths of a learning model.
To our aim, we train ten different seeded models on each class, considering the other nine as anomalies. Such a procedure allows us to estimate the mean response of our approach and its standard deviation.

Regarding the \mvtec{} and the \shanghai{} datasets, we use a residual-like structure that comprises four and five residual blocks, respectively, followed by two fully connected layers.  For this dataset, the ``selector" blocks consist of a convolutional layer followed by a pooling operation, a batch norm layer, and a final fully connected layer. In video-based AD, we substitute the ``selector" networks with LSTM cells to exploit the time correlation among the frames within a given input clip. 
To train models on those two datasets, we use again the Adam~\cite{kingma2014adam} optimizer and a learning rate in the set $\{10^{-2}, 10^{-3}\}$ that we drop by a factor of ten at specific epochs depending on the class under study. Being each class of each dataset an independent AD problem, we use different hyperparameters to train the models on each of them. Moreover, we do not always use the same set of layers to evaluate the objective in \autoref{eq:obj_full}. Indeed, we train the models by using different layer combinations and finally select the best performing one in each class. 

To allow the researchers to reproduce our work, we made the code publicly available on GitHub\footnote{\url{https://github.com/fvmassoli/mocca-anomaly-detection.git}}.

\section{Experiments} \label{exp_results}

In this section, we report our experimental results. However, before that, we describe the various metrics we use to assess the models' performance. 

\begin{table*}[!t]

\begin{tabularx}{\linewidth}{l>{\centering}X>{\centering}Xcc>{\centering}Xcc|cc}
\toprule
Class & VAE~\cite{kingma2013auto}$^\star$ & Pix CNN~\cite{van2016conditional}$^\star$ & DCAE$^\dagger$ & AnoGAN~\cite{schlegl2017unsupervised}$^{\dagger}$ & LSA~\cite{abati2019latentspace} & Deep SVDD$_{(s)}$\cite{ruff2018deep} & \MODEL{}$_{(s)}$ & Deep SVDD$_{(h)}$\cite{ruff2018deep} & \MODEL{}$_{(h)}$ \\ 
\midrule
0 & 0.688 & 0.788 & 0.601  $\pm$ .007 &  0.671   $\pm$ .025 & \b 0.735 & 0.617   $\pm$ .042 & 0.626  $\pm$ .021 & 0.617  $\pm$ .041 & 0.660  $\pm$ .015 \\

1 & 0.403 & 0.428 & 0.574  $\pm$ .029 & 0.547  $\pm$ .034 & 0.580 & 0.648  $\pm$ .014 & \b \u{0.746}  $\pm$ .008 & 0.659  $\pm$ .021 & \b 0.705  $\pm$ .013 \\

2 & 0.679 & 0.617 & 0.489  $\pm$ .024 & 0.529  $\pm$ .030 & \b 0.690 & 0.495  $\pm$ .014 & 0.575  $\pm$ .018 & 0.508  $\pm$ .008 & 0.524  $\pm$ .010 \\

3 & 0.528 & 0.574 & 0.584  $\pm$ .012 & 0.545  $\pm$ .019 & 0.542 & 0.560  $\pm$ .011 & 0.578  $\pm$ .011 & 0.591  $\pm$ .014 & \b 0.601  $\pm$ .006 \\

4 & 0.748 & 0.511 & 0.540  $\pm$ .013 & 0.651  $\pm$ .032 & \b 0.761 & 0.599  $\pm$ .011 & 0.615  $\pm$ .012 & 0.609  $\pm$ .011 & 0.609  $\pm$ .012 \\

5 & 0.519 & 0.571 & 0.622  $\pm$ .018 & 0.603  $\pm$ .018 & 0.546 & 0.621  $\pm$ .024 & \b 0.663  $\pm$ .010 & 0.657  $\pm$ .025 & \b \u{0.684}  $\pm$ .016 \\

6 & 0.695 & 0.422 & 0.512  $\pm$ .052 & 0.585  $\pm$ .014 & \b 0.751 & 0.678  $\pm$ .024 & 0.674  $\pm$ .012 & 0.677  $\pm$ .026 & 0.671  $\pm$ .005 \\

7 & 0.500 & 0.454 & 0.586 $\pm$ .029 & 0.625  $\pm$ .008 & 0.535 & 0.652  $\pm$ .010 & \b \u{0.721}  $\pm$ .004 & 0.673  $\pm$ .009 & \b 0.685  $\pm$ .010 \\

8 & 0.700 & 0.715 & 0.768  $\pm$ .014 & 0.758  $\pm$ .041 & 0.717 & 0.756  $\pm$ .017 & \b 0.791  $\pm$ .012 & 0.759  $\pm$ .012 & \b \u{0.792}  $\pm$ .008 \\

9 & 0.398 & 0.426 & 0.673  $\pm$ .030 & 0.665  $\pm$ .028 & 0.548 & 0.710  $\pm$ .011 & \b \u{0.773}  $\pm$ .010 & 0.731  $\pm$ .012 & \b 0.758  $\pm$ .007 \\



\bottomrule

\multicolumn{5}{@{}l}{$^\star$Values reported in \cite{abati2019latentspace}; $^\dagger$Values reported in \cite{ruff2018deep}}
\end{tabularx}
\caption{AUC for the \cifar{} dataset. 
The subscripts $(s)$ and $(h)$ refer to the \emph{soft} and \emph{hard} boundaries, respectively. We emphasize in bold the performance of the best models. Whenever our models overcome the SotA with both the type of boundaries, we underline the best of the two. We only report errors from others when available in the reference paper.}
\label{tab:cfr10}
\end{table*}

\subsection{Metrics}
To assess the performance of the models trained with \MODEL{} and compare them to the other approaches in the literature, we exploit two metrics: the Area Under the Curve (AUC) and the maximum Balanced Accuracy (maxBA). The former metric is the area under the Receiver Operating Characteristics curve. Instead, concerning the latter, the Balanced Accuracy (BA) represents the arithmetic mean between the sensitivity, i.e., percentage of anomalous samples correctly detected, and the specificity, i.e., same as the sensitivity but for non-anomalous samples:

\begin{flalign} \label{eq:mba}
\mathrm{BA} = \frac{TP}{2\cdot (TP\ +\ FN)} + \frac{TN}{2\cdot (TN\ +\ FP)}
\end{flalign}

where $TP$ and $FN$ are the true positives and the false negatives, respectively, and $TN$ and $FP$ are the true negatives and the false positives, respectively.

In the AD context, it is useful to quote both the AUC and the maxBA metrics.
The former one provides an aggregate measure of the performance of a model across all possible classification thresholds.
Instead, maxBA is a measure of performance at a specific threshold that could be used in production.
It selects the threshold for which the balanced accuracy measure, i.e., the average among the correctly classified images for anomalous (true positives) and anomaly-free test images (true negatives), is maximum and reports the obtained BA.
We evaluate both metrics only on the \mvtec{} dataset since for the \cifar{} and \shanghai{} datasets we only found the AUC values reported in the literature. Concerning the anomaly score for a given input image, we evaluate its value as:

\begin{flalign} \label{eq:anscore}
\tau_j(\mathbf{x}) &= \parallel \phi_j(\mathbf{x}, \theta) - \mathbf{c}_j \parallel^2 \\ \nonumber
\gamma(\mathbf{x}) &= \frac{1}{|\mathcal{J}|}\sum_j^{|\mathcal{J}|}
\begin{cases}
\tau_j(\mathbf{x}) & \text{hard boundary}\\
\\
\tau_j(\mathbf{x}) - R_j^2 & \text{soft boundary}\\
\end{cases}
\end{flalign}

where $\mathbf{x}$ is the input image, $\mathcal{J} = \{ j\ |\ j=\ 0,\ 1,\ ... J\}$ is the set of layers we consider, $\phi_j(\mathbf{x}, \theta)$ is the feature vector extracted at layer $j$, and $\mathbf{c}_j$ and $R_j$ are the center of the hypersphere and its radius at the layer $j$, respectively and $\gamma$ is the anomaly score. We refer the reader to \autoref{approach} for further details on the meaning of the boundaries. Concerning the textures-type classes from the \mvtec{}, we evaluate the anomaly score as the maximum among the scores relative to each of the 64x64 patches of the given image:

\begin{flalign} \label{eq:mvtec_pathces_score}
\gamma^{h/s}(\mathbf{x}) = \mathrm{max}\big\{\gamma^{h/s}(\mathrm{patch}_i))\ \ |\ \ i\ =\ 1,\ 2,\ ...,\ 64 \big\}
\end{flalign}

where the superscripts $s$ and $h$ correspond to when we apply a ``soft" or ``hard" boundary while training the model, respectively. More details on how we extract patches from a single image can be found in \autoref{mvtecpreprocess}. Lastly, considering video-based input we consider a single input clip as made of 16 frames. We then apply a sliding window technique to move through all the frames of a given video and construct the input clips. Since each frame can appear multiple times across different clips, we evaluate its score as the mean value among all of its scores. Moreover, a single frame can have different scores in different clips having a different time correlation, captured by the LSTMs (see \autoref{fig:arch}), with all the other frames. For such a reason, we normalize the score of each frame to the maximum and minimum values of the scores within the clips in which the frame under analysis is present:

\begin{flalign} \label{eq:videoan}
\gamma^{h/s}(\mathbf{x}_i) = \frac{\langle \gamma^{h/s}(\mathbf{x}_i) \rangle - \mathrm{max_{clips}}\langle \gamma^{h/s}(\mathbf{x}_i) \rangle}{\mathrm{max_{clips}}\langle \gamma^{h/s}(\mathbf{x}_i) \rangle - \mathrm{min_{clips}}\langle \gamma^{h/s}(\mathbf{x}_i) \rangle}
\end{flalign}

Finally, we add a reconstruction term to the score.

\subsection{Experimental results - CIFAR10}

Concerning the \cifar{} dataset, we instantiate each class as a single AD problem, and we train ten different seeded models on each of them. Such a procedure allows us to quote a mean AUC value and the corresponding variance. We report the results in \autoref{tab:cfr10}.

As we can see from \autoref{tab:cfr10}, our approach reaches the highest performance on six out of ten classes. Moreover, on class-1, class-5, class-7, class-8, and class-9, the \MODEL{} method performs better than the state-of-the-art (SotA) results concerning both the ``soft" and the ``hard" boundaries. As reported in \autoref{train_details}, on the \cifar{} dataset we use a LeNet-like architecture as in~\cite{ruff2018deep}. 
Moreover, to better emphasize that our approach's higher performance is not due to a mere addition of more models to the baseline, we use averaging pooling layers as ``selectors" blocks.
Thus, since we use the same architecture as in~\cite{ruff2018deep}, we can conclude that the higher performance of our models are only due to the use of \MODEL{} and not because we use deeper models or because we add more branches to the base architecture.
To summarize the previous results, we report in \autoref{tab:cfr10_means} the AUC values, for each model in \autoref{tab:cfr10}, averaged among all the ten classes of the dataset.

\begin{table}[!ht]

\begin{tabularx}{\linewidth}{l>{\centering\arraybackslash}X}
\toprule
& Average AUC \\
\cmidrule{2-2}
VAE~\cite{kingma2013auto}$^\star$ & 0.586 $\pm$ .039 \\
Pix CNN~\cite{van2016conditional}$^\star$ & 0.551 $\pm$ .038 \\
DCAE$^\dagger$ & 0.595 $\pm$ .024 \\
AnoGAN~\cite{schlegl2017unsupervised}$^{\dagger}$ & 0.618 $\pm$ .021 \\
LSA~\cite{abati2019latentspace} & 0.640 $\pm$ .029 \\
Deep SVDD$_{(s)}$\cite{ruff2018deep} & 0.634 $\pm$ .022 \\
Deep SVDD$_{(h)}$\cite{ruff2018deep} & 0.648 $\pm$ .022 \\
\cmidrule{2-2}
\MODEL{}$_{(s)}$ & \b \u{0.676} $\pm$ .024 \\
\MODEL{}$_{(h)}$ & \b 0.669 $\pm$ .023\\ 

\bottomrule

\multicolumn{2}{@{}l}{$^\star$Values reported in \cite{abati2019latentspace}; $^\dagger$Values reported in \cite{ruff2018deep}}

\end{tabularx}

\caption{AUC averaged among all classes of the \cifar{} dataset. The subscripts $(s)$ and $(h)$ refer to the \emph{soft} and \emph{hard} boundaries, respectively. We emphasize in bold the performance of the best models. Whenever our models overcome the SotA with both the type of boundaries, we underline the best of the two.}
\label{tab:cfr10_means}
\end{table}

From \autoref{tab:cfr10_means}, it is clear that our approach reaches the highest performance concerning both types of boundary settings. Moreover, we can appreciate that we obtain higher performance, also considering larger models such as LSA~\cite{abati2019latentspace}.

\subsection{Experimental results - MVTec AD}

\begin{table*}[!t]
\setlength{\tabcolsep}{2.5pt}
\centering
\begin{tabularx}{\linewidth}{l>{\centering}X>{\centering}X>{\centering}X>{\centering}X>{\centering}X>{\centering}X>{\centering}X>{\centering}X>{\centering}X>{\centering}X>{\centering}X>{\centering}Xcc>{\centering\arraybackslash}X}

\toprule
& \multicolumn{5}{c}{Textures} & \multicolumn{10}{c}{Objects} \\

\cmidrule{2-6} \cmidrule(l){7-16}

& \m{Carpet} & \m{Grid} & \m{Leather} & \m{Tile} & \m{Wood} & \m{Bottle} & \m{Cable} & \m{Capsule} & \m{Hazelnut} & \m{MetalNut} & \m{Pill} & \m{Screw} & \m{Toothbrush} & \m{Transistor} & \m{Zipper}  \\
       
\cmidrule{2-6} \cmidrule(l){7-16}

AE$_\text{SSIM}$~\cite{bergmann2018improving}$^\star$     &     0.67 &     0.69 &     0.46 &     0.52 &     0.83 &  0.88 &     0.61 &    0.61 &    0.54 &     0.54 &    0.60 &    0.51 &     0.74 &    0.52 &    0.80  \\

AE$_\text{L2}$~\cite{bergmann2018improving}$^\star$       &     0.50 &     0.78 &     0.44 &     0.77 &     0.74  &     0.80 &     0.56 &    0.62 & \b 0.88 &     0.73 &    0.62 &    0.69 &\b 0.98 &    0.71 &    0.80 \\

AnoGAN~\cite{schlegl2017unsupervised}$^\dagger$           &     0.49 &     0.51 &     0.52 &     0.51 &     0.68 &     0.69 &     0.53 &    0.58 &    0.50 &     0.50 &    0.62 &    0.35 &     0.57 &    0.67 &    0.59 \\
VAE-grad~\cite{dehaene2020iterative}$^\dagger$            &     0.67 &     0.83 &     0.71 &     0.81 & 0.89 &  0.86 &     0.56 &\b  0.86 &    0.74 & 0.78 &    0.80 &    0.71 &     0.89 &    0.70 &    0.67 \\

AVID~\cite{sabokrou2018avid}$^\dagger$                    &     0.70 &     0.59 &     0.58 &     0.66 &     0.83 &  0.88 &     0.64 &    0.85 &    0.86 &     0.63 &\b  0.86 &    0.66 &     0.73 &    0.58 &    \b0.84  \\

EGBAD~\cite{zenati2018efficient}$^\ddagger$                          &     0.60 &     0.50 &     0.65 &     0.73 &     0.80 & 0.68 &     0.66 &    0.55 &    0.50 &     0.55 &    0.63 &    0.50 &     0.48 &    0.68 &    0.59 \\

CBiGAN~\cite{carrara2020combining} &     0.60 &\b 0.99 & 0.87 & \b 0.84 &     0.88 &  0.84 &\b 0.73 &    0.58 &    0.75 &     0.67 &    0.76 &    0.67 &     0.97 &  0.74 &    0.55  \\

\cmidrule{2-6} \cmidrule(l){7-16}

\MODEL{}$_{(s)}$ & \b \u{0.81} & 0.85 & \b \u{0.96} & 0.80 & \b \u{0.97} & \b \u{0.90} & 0.72 & 0.77 & 0.77 & \b \u{0.85} & 0.81 & \b\u{0.82} & 0.93 & \b0.77 & 0.78  \\

\MODEL{}$_{(h)}$ & \b 0.74 & 0.76 & \b 0.91 & 0.78 & \b 0.94 & \b \u{0.90} & 0.68 & 0.75 & 0.76 & \b 0.80 & 0.69 & \b0.80 & 0.91 & \b\u{0.81} &  0.78 \\

\bottomrule

\multicolumn{10}{@{}l}{$^\star$Values reported in \cite{bergmann2019mvtec}; $^\dagger$Values reported in \cite{venkataramanan2019attention}; $^\ddagger$Values reported in \cite{carrara2020combining}}

\end{tabularx}

\caption{maxBA for all the classes of the \mvtec{} dataset. The subscripts $(s)$ and $(h)$ refer to the \emph{soft} and \emph{hard} boundaries, respectively. We emphasize in bold the performance of the best models. Whenever our models overcome the SotA with both the type of boundaries, we underline the best of the two.}
\label{tab:mvtec_balacc}
\end{table*}

\begin{table*}[!t]
\setlength{\tabcolsep}{2.5pt}
\centering
\begin{tabularx}{\linewidth}{l>{\centering}X>{\centering}X>{\centering}X>{\centering}X>{\centering}X>{\centering}X>{\centering}X>{\centering}X>{\centering}X>{\centering}X>{\centering}X>{\centering}Xcc>{\centering\arraybackslash}X}
\toprule

& \multicolumn{5}{c}{Textures} & \multicolumn{10}{c}{Objects} \\

\cmidrule{2-6} \cmidrule(l){7-16}

& \m{Carpet} & \m{Grid} & \m{Leather} & \m{Tile} & \m{Wood} & \m{Bottle} & \m{Cable} & \m{Capsule} & \m{Hazelnut} & \m{MetalNut} & \m{Pill} & \m{Screw} & \m{Toothbrush} & \m{Transistor} & \m{Zipper}  \\

\cmidrule{2-6} \cmidrule(l){7-16}

AE$_\text{L2}$~\cite{bergmann2018improving}$^\dagger$  &    0.64 &    0.83 &    0.80 &    0.74 & 0.97 &  0.65 &    0.64 &    0.62 &    0.73 &    0.64 &    0.77 & \b 1.00 &    0.77 &    0.65 &    0.87 \\

GeoTrans~\cite{golan2018deep}$^\dagger$    &    0.44 &    0.62 &    0.84 &    0.42 &    0.61 &   0.74 &    0.78 &    0.67 &    0.36 & 0.81 &    0.63 &    0.50 &    0.97 &  0.87 &    0.82  \\

GANomaly~\cite{akcay2018ganomaly}$^\dagger$&    0.70 &    0.71 &    0.84 &    0.79 &    0.83 &  0.89 &    0.76 & 0.73 &    0.79 &    0.70 &    0.74 &    0.75 &    0.65 &    0.79 &    0.75\\

ITAE~\cite{huang2019inverse}     & 0.71 &    0.88 & 0.86 &    0.74 &    0.92 & 0.94 & \b 0.83 &    0.68 & \b 0.86 &    0.67 &    0.79 & \b 1.00 & \b 1.00 &    0.84 & \b 0.88 \\

EGBAD~\cite{zenati2018efficient}$^\star$ &    0.52 &    0.54 &    0.55 &    0.79 &    0.91 &    0.63 &    0.68 &    0.52 &    0.43 &    0.47 &    0.57 &    0.46 &    0.64 &    0.73 &    0.58  \\

CBiGAN~\cite{carrara2020combining} &    0.55 & \b 0.99 &    0.83 & \b 0.91 &    0.95 & 0.87 &    0.81 &    0.56 &    0.77 &    0.63 & 0.81 &    0.58 &    0.94 &    0.77 &    0.53  \\

CAVGA-R$_{u}$~\cite{venkataramanan2020attention}$^{\star\star}$ & \it 0.73 & \it 0.75 & \it 0.71 & \it 0.70 & \it 0.85 & \it 0.89 & \it  0.63 & \it 0.83 & \it 0.84 & \it 0.67 & \it 0.88  & \it 0.77 & \it 0.91 & \it 0.73 & \it 0.87\\
CAVGA-D$_{u}$~\cite{venkataramanan2020attention}$^{\star\star}$ & \it 0.78 & \it 0.78 & \it 0.75 & \it 0.72 & \it  0.88 & \it 0.91 & \it 0.67 & \it 0.87 & \it 0.87 & \it 0.71 & \it 0.91 & \it 0.78 & \it 0.97 & \it 0.75 & \it 0.94\\

\cmidrule{2-6} \cmidrule(l){7-16}

\MODEL{}$_{(s)}$ & \b \u{0.86} & 0.87 & \b \u{0.98} & 0.89 & \b \u{1.00} & \b 0.95 & 0.76 & \b \u{0.82} & 0.80 & \b \u{0.85} & \b 0.82 & 0.84 & 0.97 & \b0.88 & 0.84 \\

\MODEL{}$_{(h)}$ & \b 0.74 & 0.81 & \b 0.95 & 0.85 & \b 0.97 & 0.93 & 0.72 & \b 0.79 & 0.78 & \b 0.84 & 0.73 & 0.80 & 0.95 & 0.84 & 0.82 \\

\bottomrule

\multicolumn{16}{@{}l}{$^\dagger$Values reported in \cite{huang2019inverse}; $^\star$Values reported in \cite{carrara2020combining}; $^{\star\star}$Results obtained by using more data - should NOT be directly compared to all the other methods} \\ 

\end{tabularx}

\caption{AUC for all the classes of the \mvtec{} dataset. The subscripts $(s)$ and $(h)$ refer to the \emph{soft} and \emph{hard} boundaries, respectively. We emphasize in bold the performance of the best models. Whenever our models overcome the SotA with both the type of boundaries, we underline the best of the two.}
\label{tab:mvtec_auroc}
\end{table*}

Regarding the \mvtec{} dataset, also, in this case, we consider each class as an independent AD problem. As reported in \autoref{mvtecpreprocess}, the dataset classes are divided into texture- and object-like sets. For each class, we report the maxBA and the AUC in \autoref{tab:mvtec_balacc} and \autoref{tab:mvtec_auroc}, respectively.
Regarding the texture-type of classes, we see from the tables that the \MODEL{} approach allows our models to reach the highest performance on three out of five classes concerning both the \emph{hard} and \emph{soft} boundaries. Similar reasonings hold in the case of object-type classes, too. 
Concerning the results from~\cite{venkataramanan2020attention}, it is important to highlight that, even though we report their results, they should not directly compared with others. The reason for that is because in~\cite{venkataramanan2020attention}, the models are trained on more data rather than on MVTec AD only. Thus, those results are not directly comparable with the other methods.
Due to the very low number of test images available in the dataset, typically large variations in the performance of the model are observed among the different classes. Thus, to better compare the performance of the various approaches, we report in \autoref{tab:mvtec_means} the overall mean values for the maxBA and the AUC evaluated among all classes of the dataset.


\begin{table}[!ht]
\begin{tabularx}{\linewidth}{l>{\centering}X>{\centering\arraybackslash}X}
\toprule

& \multicolumn{2}{c}{Overall Mean} \\
\cmidrule{2-3}
 & maxBA & AUC  \\ 
\cmidrule{2-3}

AE$_\text{SSIM}$~\cite{bergmann2018improving}$^\star$ & 0.63 & \nasymb{}  \\
AE$_\text{L2}$~\cite{bergmann2018improving}$^\star$ & 0.71 & 0.75 \\
AnoGAN~\cite{schlegl2017unsupervised}$^\dagger$ & 0.55 & \nasymb{} \\
VAE-grad~\cite{dehaene2020iterative}$^\dagger$  & 0.77 & \nasymb{} \\
AVID~\cite{sabokrou2018avid}$^\dagger$ & 0.73 & \nasymb{} \\
EGBAD~\cite{zenati2018efficient}$^{\ddagger}$  & 0.61 & 0.60 \\
GeoTrans~\cite{golan2018deep}$^{\dagger\dagger}$ & \nasymb{} & 0.67 \\
GANomaly~\cite{akcay2018ganomaly}$^{\dagger\dagger}$ & \nasymb{} & 0.76 \\
ITAE~\cite{huang2019inverse}$^{\dagger\dagger}$ &\nasymb{} & 0.84 \\
CBiGAN~\cite{carrara2020combining} & 0.76 & 0.77 \\
\cmidrule{2-3}

\MODEL{}$_{(s)}$ &  \b \u{0.83} & \b 0.88 \\
\MODEL{}$_{(h)}$ & \b 0.80 & 0.83 \\
\bottomrule

\multicolumn{3}{@{}l}{$^\star$Values reported in \cite{bergmann2019mvtec}; $^\dagger$Values reported in \cite{venkataramanan2019attention}} \\

\multicolumn{3}{@{}l}{$^\ddagger$Values reported in \cite{carrara2020combining}; $^{\dagger\dagger}$Values reported in \cite{huang2019inverse}} 


\end{tabularx}

\caption{Average maxBA and AUC from \autoref{tab:mvtec_balacc} and \autoref{tab:mvtec_auroc}. The subscripts $(s)$ and $(h)$ refer to the \emph{soft} and \emph{hard} boundaries, respectively. We emphasize in bold the performance of the best models. Whenever our models overcome the SotA with both the type of boundaries, we underline the best of the two. The ``\nasymb{}" symbol means that the authors did not report the value.}
\label{tab:mvtec_means}

\end{table}

From the table, we conclude that the \MODEL{} approach allows us to reach the highest performance on both types of metrics considering both the \emph{soft} and \emph{hard} type of boundary.

\subsection{Experimental results - ShanghaiTech}

Differently from the \cifar{} and \mvtec{} datasets, the \shanghai{} concerns the video-based AD task. Although we test models trained with \MODEL{} against such a protocol, it is essential to stress that our approach is not specially designed for the video-based scenario. We report our results in \autoref{tab:shanghai} and others available in the literature. 

\begin{table}[!ht]
\begin{tabularx}{\linewidth}{l>{\centering\arraybackslash}X}
\toprule
 & AUC \\ 

\cmidrule{2-2}

AE-Conv2D~\cite{hasan2016learning}$^\dagger$  & 0.609   \\
TSC~\cite{luo2017stackrnn}$^\dagger$  & 0.679  \\
Stack RNN~\cite{luo2017stackrnn}$^\dagger$ & 0.680   \\
AE-Conv3D~\cite{zhao2017spatiotemporal}$^\dagger$  & 0.697   \\
MemAE~\cite{gong2019memorizing}$^\dagger$  & 0.712 \\
LSA~\cite{abati2019latentspace} & 0.725  \\
ITAE~\cite{huang2019inverse} & 0.725 \\
FFP+MC~\cite{liu2018futureframe} & 0.728  \\
Mem-Guided (w/o Mem.)~\cite{park2020learning} & 0.668 \\
Mem-Guided (w/ Mem.)~\cite{park2020learning} & 0.705 \\
MemAE-nonSpar~\cite{gong2019memorizing} & 0.688 \\
MemAE~\cite{gong2019memorizing} & 0.712 \\
Clustering-Driven~\cite{chang2020clustering} & \b 0.733 \\

\cmidrule{2-2}

\MODEL{}$_{(s)}$ & 0.730 \\ 
\MODEL{}$_{(h)}$ & 0.725 \\ 

\bottomrule

\multicolumn{1}{@{}l}{$^\dagger$Values reported in \cite{huang2019inverse}}
\end{tabularx}
\caption{AUC values for the ShanghaiTech~\cite{luo2017revisit} dataset. The subscripts $(s)$ and $(h)$ refer to the \emph{soft} and \emph{hard} boundaries, respectively. We report in bold the performance of the best model.}
\label{tab:shanghai}
\end{table}

From \autoref{tab:shanghai}, we can see that our approach's performance is utterly comparable to the current SotA models, specifically designed to handle video-based input. Thus, showing that our method is applicable to both the image- and video-based anomaly detection tasks. Indeed, the only modification we apply to \MODEL{} for video-based contexts is to move to a single-step training, based on the same objectives, and to substistute the ``selector" modules with LSTMs. 
\section{Model Analysis}\label{model_analysis}

In this section, we look in more detail at the behavior of our models. First, we focus on an ablation study to show the impact of using a different number of layers to evaluate a specific image's anomaly score. Specifically, we prove that with \MODEL{}, we effectively succeed in exploiting the deep representations extracted at different depths of a DNN. To our aim, we perform the ablation study considering the ``Leather" class of the \mvtec{} dataset. We report the results in \autoref{tab:ablation_mvtec}. 

\begin{table}[!h]
\begin{tabularx}{\linewidth}{l>{\centering}X>{\centering\arraybackslash}X}
\toprule
Layer index & \multicolumn{2}{c}{maxBA} \\ 
\cmidrule{2-3}
& hard boundary & soft boundary \\
\cmidrule{2-3}
6 & 0.819 $\pm$ .020 & 0.839 $\pm$ .010 \\ 
5, 6  & 0.855 $\pm$ .021 & 0.840 $\pm$ .012 \\ 
4, 5, 6 & 0.906 $\pm$ .001 & \b 0.955 $\pm$ .007 \\ 
3, 4, 5, 6  & \b 0.912 $\pm$ .002 & 0.935 $\pm$ .005 \\ 
2, 3, 4, 5, 6 & 0.903 $\pm$ .003 & 0.947 $\pm$ .005 \\ 
1, 2, 3, 4, 5, 6 & 0.865 $\pm$ .004 & 0.948 $\pm$ .003 \\
0, 1, 2, 3, 4, 5, 6 & 0.873 $\pm$ .002 & 0.924 $\pm$ .001 \\
\bottomrule
\end{tabularx}
\caption{Ablation study concerning the ``Leather" class of the MVTec AD~\cite{bergmann2019mvtec} dataset. We report the maxBA for the \emph{hard} and \emph{soft} boundary settings. We highlight in bold the best results.} 
\label{tab:ablation_mvtec}
\end{table}

As described in \autoref{train_details}, the encoder's architecture consists of four residual blocks followed by two fully connected layers.
The indexes in the first column of \autoref{tab:ablation_mvtec} correspond to the layers' ordering where the $0$-th layer is the closest to the input. The results in \autoref{tab:ablation_mvtec} should be interpreted as follows. Each row in the table represents a different model that we trained with \MODEL{} by considering the output from the layers listed in the first column. For example, the first row represents the results we obtained by considering the output (in the training and test phases) from the last layer only, while in the second row we consider the layer 5 and 6 together. We aim at showing that by exploiting the output at different layers while training a learning model, we can use the output from those same layers at inference time to enhance the network's discrimination power. On the contrary, we experimentally observed that training the model using the last layer's output only and then using more layers at inference time always gave worse results. Such an observation is one of the key points on which we base our approach.

As it is clear from the table, independently from the type of boundary we apply, we obtain higher results by utilizing more layers. This result supports our intuition that the features extracted at different depths help to detect anomalies in the input images.
By carefully looking at \autoref{tab:ablation_mvtec} we notice that the maxBA improves until we add layers 4 and 5 to the last one. Moreover, we can notice that, in the case of the \emph{hard} boundary setting, we can obtain a slight improvement by adding layer 3. Finally, we notice that by adding more layers, we do not see any further improvement. We can interpret such behavior by considering that since the first layers are closer to the input data, they specialize on simple patterns. On the contrary, higher layers generate representations that amplify aspects of the input that are important for discrimination~\cite{lecun2015deep}, thus more useful to fulfill the final task. Hence, adding layers that are too close to the input data does not improve the learning model's overall performance. 

\begin{figure}[!h]
\includegraphics[width=\linewidth]{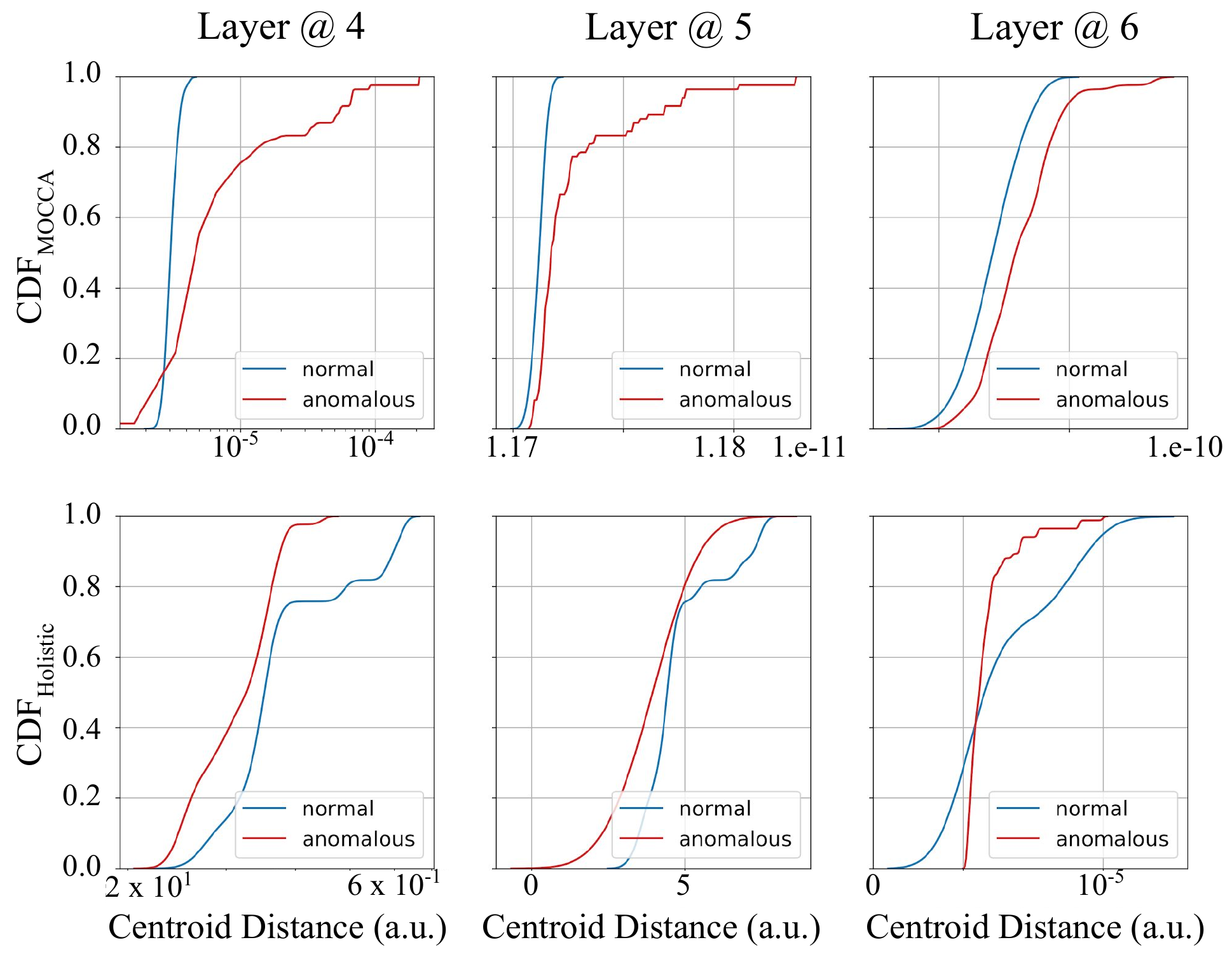}
\caption{CDF of the test images distance from the ``normal" class centroid for the \MODEL{} approach (top row) and for the ``holistic" one (bottom row). The blue (red) line represents the CDF of ``normal" (anomalous) images.} \label{fig:mvtec_cdf}
\end{figure}

We then focus our attention on the distribution of the distances among the features and the centroids, of a given class, at different layers. Specifically, we compare our training approach against the ``holistic" approach, i.e., when the learning model is considered a single computational block. 
As specified in \autoref{introduction}, ``holistic" refers to the approach similar to~\cite{ruff2018deep} where the last layer's output only is used to train and test the encoder on the AD task.
To our aim, we train two identical models once with the \MODEL{} approach and then by evaluating the OC loss on the last layer only (``holistic"). We report the resulting Cumulative Density Function (CDF) in \autoref{fig:mvtec_cdf}.

Concerning the model trained with \MODEL{}, we see that the distributions for ``normal" images always lies at the left of the corresponding for anomalous ones (as one would expect). Moreover, we see that the CDFs of ``normal" images rise faster than the ones of anomalous samples. Thus, allowing one to set a more discriminative threshold on the anomaly score. On the contrary, we see that by considering the last layer only while training on the AD task, the distributions of distances for anomalous and ``normal" images are highly overlapped even in the last layer. Thus, by training with \MODEL{}, we have a double gain: on the one side, we obtain discriminative deep features from more layers, and on the other hand, we are able to set more discriminative thresholds. 

\section{Conclusions} \label{conclusions}

The anomaly detection task is still an open challenge in many scientific fields. Several approaches have been proposed to tackle this problem in the context of deep learning, typically based on an unsupervised training paradigm. Indeed, being rare events, collecting anomalous samples to construct a supervised training dataset might be extremely expensive. Thus, approaches in which neural networks automatically learn the concept of ``normality" from non-anomalous data only represent a promising solution. 

We propose to adopt a multi-layer approach, named \MODEL{}, to exploit the output of a deep model at different depths to detect anomalous input in the one-class setting. 
Differently from the usual ``holistic" interpretation of a learning model in which a neural network is considered a single computational block, \MODEL{} explicitly leverages the networks' multi-layer composition. Specifically, we show that such an approach enhances a neural network's discrimination capability. 
We conduct extensive experiments on three different datasets and perform an analysis of the models to support our intuitions. We test our method against the single-image AD task showing that it improves the state-of-the-art both on the CIFAR10 and MVTec AD datasets. Specifically, concerning the performance averaged among all the classes, \MODEL{} improves upon the literature results with both the \emph{soft} and \emph{hard} type of boundary. We acknowledge the best improvement concerning the overall maxBA on the MVTec AD dataset that overcomes the state-of-the-art results by 6\%. 
Moreover, even though our approach is not tailored for the video-based AD task, we test it also using such a protocol by employing the ShanghaiTech dataset. From the experimental results, we see that with \MODEL{},  the models' performance is utterly comparable to what was obtained by approaches specially designed for such a task. Thus, showing the high generalization capability of our method.

Finally, we report insights about the behavior of models trained with \MODEL{} by performing an ablation study and reporting the different CDFs of the distance of the deep representations from the centroids of a given class across different layers. Such an analysis, pointed out that the benefits from using \MODEL{} are two-fold: on the one side, we obtain discriminative deep features from more layers, and on the other hand, we are able to set more discriminative thresholds. 

%
\IEEEpeerreviewmaketitle

\section*{Acknowledgments}
We gratefully acknowledge the support of NVIDIA Corporation with the donation of the Titan V GPU used for this research.
This  work  was  partially  supported  by WAC@Lucca funded by Fondazione Cassa di Risparmio di Lucca,
AI4EU - an EC H2020 project (Contract  n.  825619), 
and upon work from COST Action 16101 ``Action MULTI-modal Imaging of FOREnsic SciEnce Evidence (MULTI-FORESEE)'', supported by COST (European Cooperation in Science and Technology).

\ifCLASSOPTIONcaptionsoff
  \newpage
\fi



%

{\small
\bibliographystyle{IEEEtran}
\bibliography{bibl}
}

%

\begin{IEEEbiography}[{\includegraphics[width=1in,height=1.25in,clip,keepaspectratio]{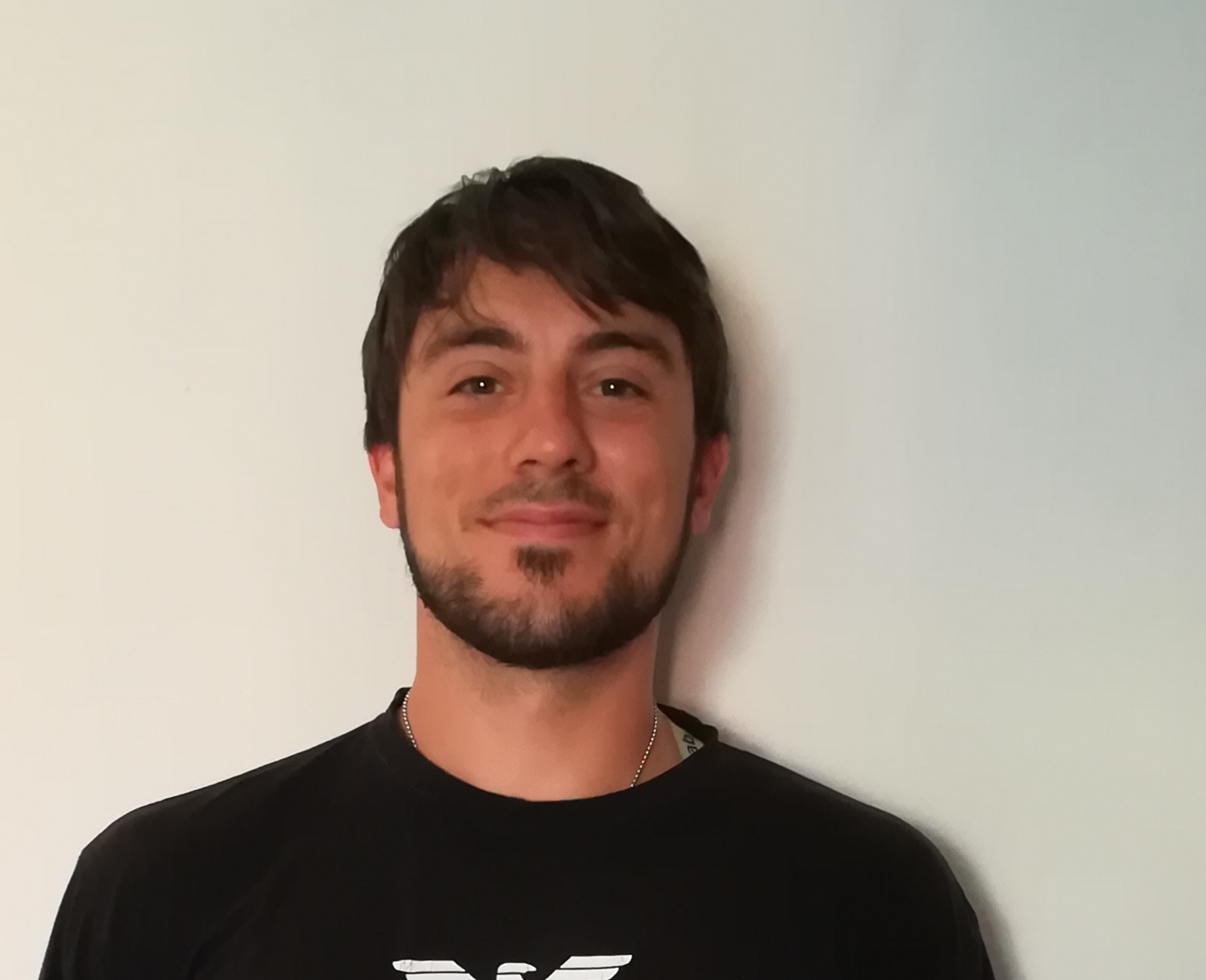}}]{Dr. Fabio Valerio Massoli}
is a PostDoc at the Artificial Intelligence for Media and Humanities lab of ISTI-CNR. He has a Ph.D. in High Energy Physics from University of Bologna in collaboration with the Columbia University (NY). 
His research interests include deep learning, supervised and unsupervised learning, generative models, and quantum theory and technologies.
\end{IEEEbiography}
\begin{IEEEbiography}[{\includegraphics[width=1in,height=1.25in,clip,keepaspectratio]{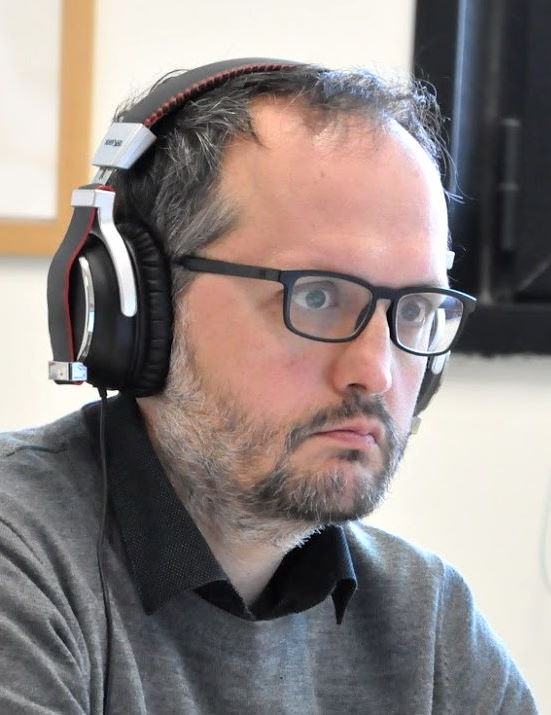}}]{Dr. Fabrizio Falchi}
is researcher of the Artificial Intelligence for Media and Humanities lab of ISTI-CNR. He has a Ph.D. in Information Engineering from University of Pisa, and a Ph.D. in Informatics from Faculty of Informatics of Masaryk Univ. of Brno. He also received an M.B.A. from Scuola Superiore Sant'Anna in Pisa.
His research interests include deep learning, convolutional neural network, similarity search, distributed indexes, multimedia information retrieval, computer vision.
\end{IEEEbiography}
\begin{IEEEbiography}[{\includegraphics[width=1in,height=1.25in,clip,keepaspectratio]{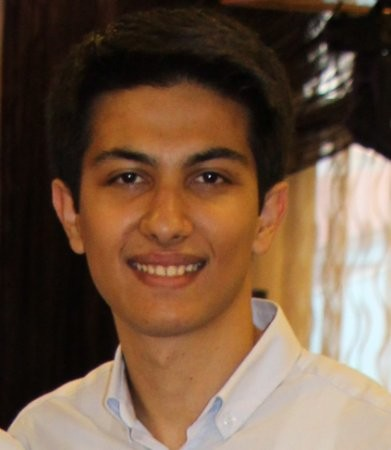}}]{Alperen Kantarci}
received his B.S. degree in Computer Engineering at Istanbul Technical University in 2019. He is currently pursuing M.Sc. degree in Computer Engineering at Istanbul Technical University. His research interests include computer vision, deep learning, contrastive learning and unsupervised learning. 
\end{IEEEbiography}
\begin{IEEEbiography}[{\includegraphics[width=1in,height=1.25in,clip,keepaspectratio]{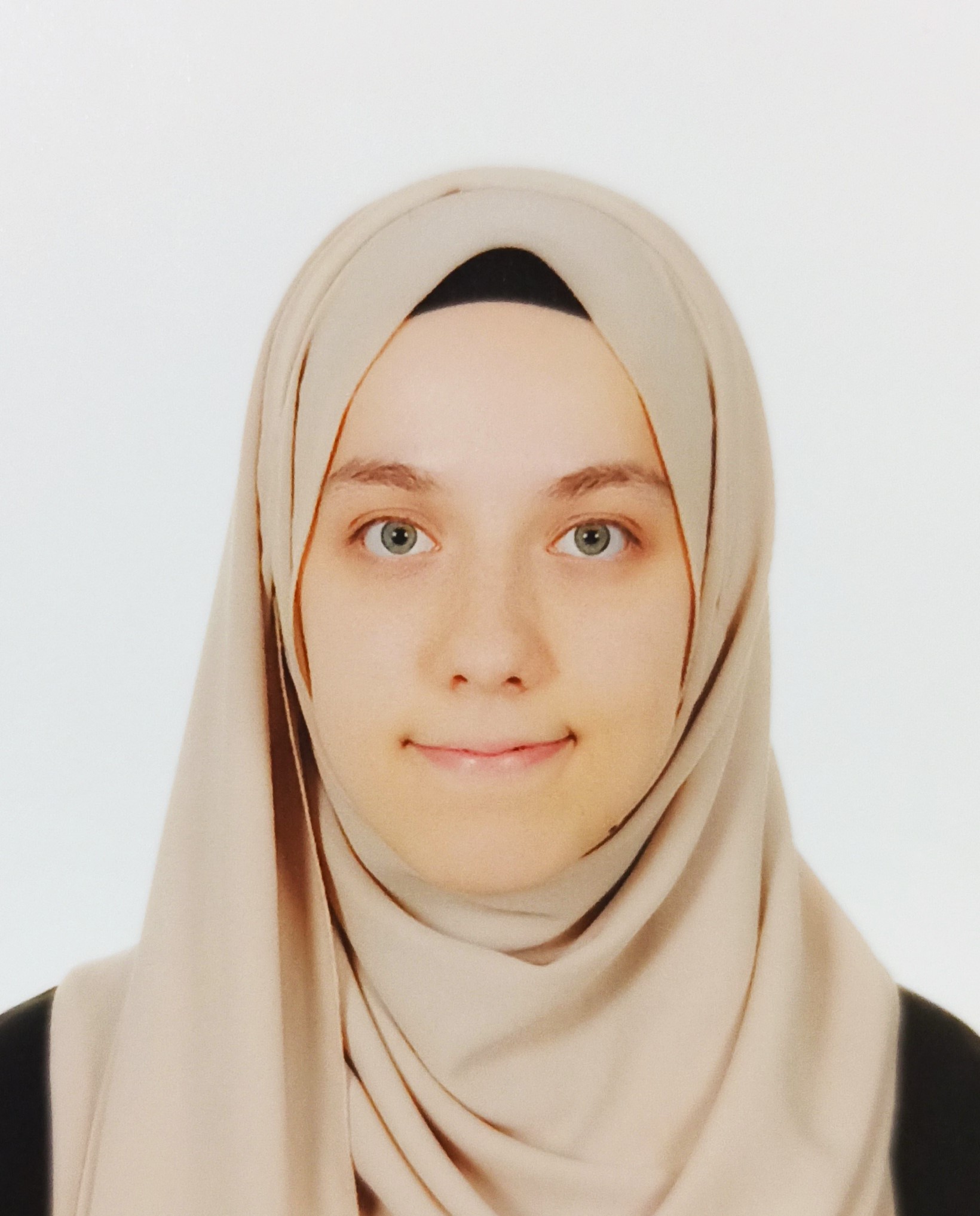}}]{\c{S}eymanur Akti}
is a M.Sc. student and research assistant at department of computer engineering in Istanbul Technical University. She has received her B.S. degree in computer engineering from Istanbul Technical University in 2019. Her research interests include deep learning, computer vision, imbalanced data classification and anomaly detection. 
\end{IEEEbiography}
\begin{IEEEbiography}[{\includegraphics[width=1in,height=1.25in,clip,keepaspectratio]{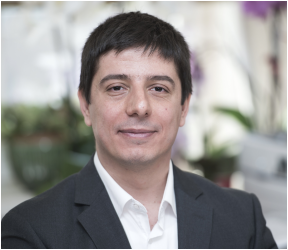}}]{Dr. Hazim Kemal Ekenel}
is a Professor at the Department of Computer Engineering in Istanbul Technical University. He received his PhD degree in Computer Science from the University of Karlsruhe (TH) in 2009. His research interest covers computer vision and machine learning with a focus on face analysis. 
He is a recipient of the Science Academy Turkey's Young Scientist Award 2018 and IEEE Turkey Section's Research Award 2019.
\end{IEEEbiography}
\begin{IEEEbiography}[{\includegraphics[width=1in,height=1.25in,clip,keepaspectratio]{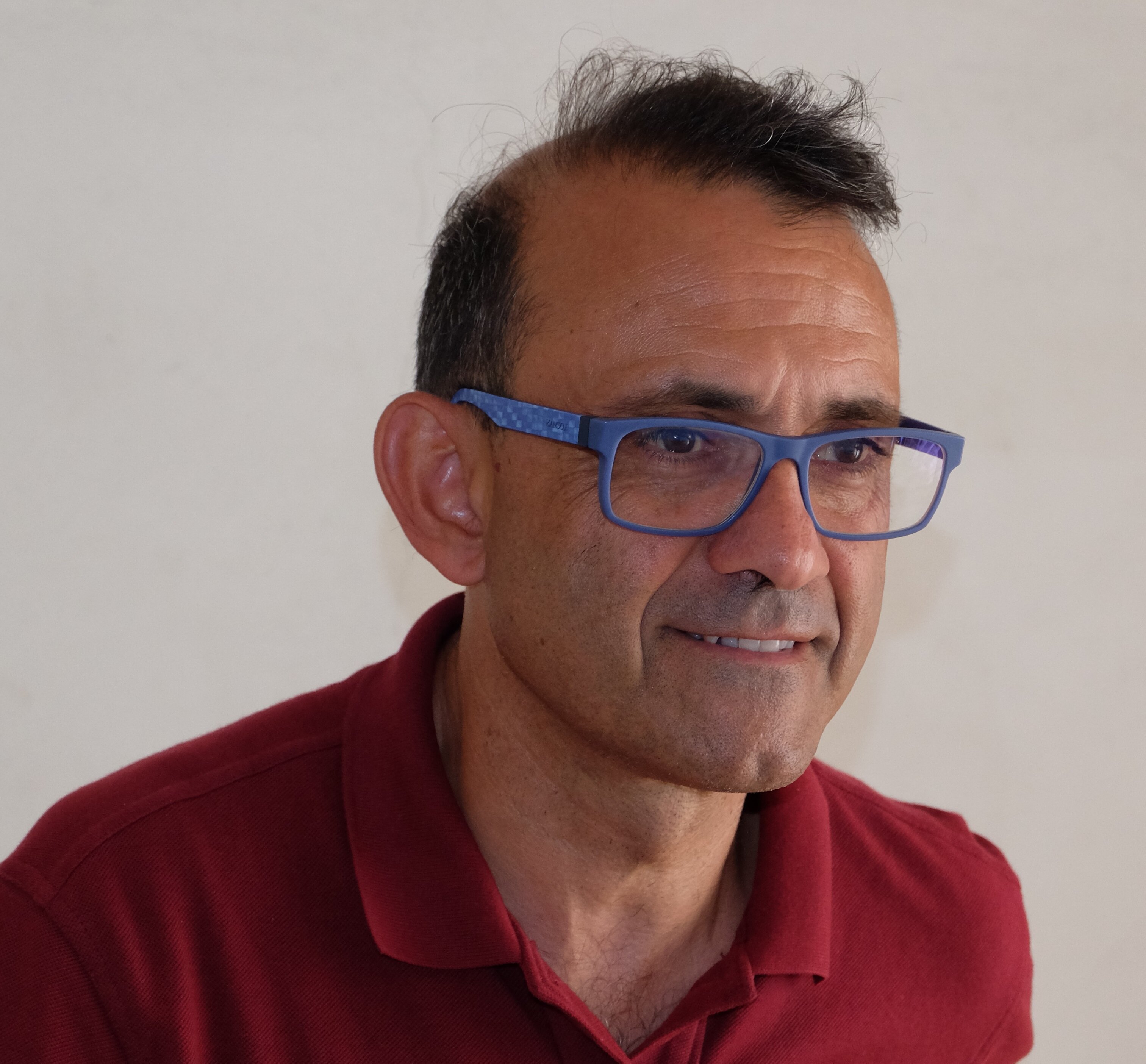}}]{Dr. Giuseppe Amato}
was awarded a PhD in Computer Science at the University of Dortmund, Germany, in 2002. He is a senior researcher at CNR-ISTI in Pisa, where he leads the ``Artificial Intelligence for Multimedia and Humanities" (AIMH) laboratory. His main research interests are artificial intelligence, content-based retrieval of multimedia documents, access methods for similarity search.
\end{IEEEbiography}



\end{document}